\pdfoutput=1

\documentclass[11pt]{article}
\usepackage{color}
\usepackage{times}
\usepackage{epsfig}
\usepackage{graphicx}
\usepackage{amsmath}
\usepackage{amsthm}
\usepackage{amssymb}
\usepackage{mathtools}
\usepackage{multirow}
\usepackage{bbm}
\usepackage{dsfont}

\usepackage[table, dvipsnames]{xcolor}

\usepackage[utf8]{inputenc}
\usepackage{booktabs,siunitx}
\usepackage{multirow}
\usepackage{multicol}
\usepackage{amsmath}
\usepackage{subcaption}
\usepackage{caption}
\usepackage{graphicx}
\usepackage{pifont}

\usepackage{tikz}
\usetikzlibrary{shapes,arrows,patterns}
\usetikzlibrary{positioning}
\usetikzlibrary{calc}

\newcommand{\cut}[1]{}

\newcommand{\minipostspace}{\vskip -2mm}

\definecolor{red}{RGB}{255, 117, 115}
\definecolor{green}{RGB}{171, 255, 175}

\usepackage{EMNLP2023}

\usepackage{times}
\usepackage{latexsym}

\usepackage[T1]{fontenc}

\usepackage[utf8]{inputenc}

\usepackage{microtype}

\usepackage{inconsolata}

\newcommand{\gray}{\cellcolor{red!40}}
\newcommand{\red}{\cellcolor{red!40}}
\newcommand{\green}{\cellcolor{green!90!black!80!white}}

\newcommand*{\relfont}{\fontfamily{bookman}\selectfont}

\title{Measuring and Modifying Factual Knowledge in Large Language Models}

\author{Pouya Pezeshkpour \\
  Megagon Labs \\
  \texttt{pouya@megagon.ai} \\
}
\begin{document}
\maketitle
\begin{abstract}
Large Language Models (LLMs) store an extensive amount of factual knowledge obtained from vast collections of text. 
To effectively utilize these models for downstream tasks, it is crucial to have reliable methods for measuring their knowledge. 
However, existing approaches for knowledge measurement have certain limitations, and despite recent efforts, they fail to provide accurate measurements and the necessary insights for modifying the knowledge within LLMs. 
In this work, we employ information theory-based measurements to provide a framework estimating the factual knowledge contained within large language models. 
More specifically, we measure knowledge by analyzing the LLM's prediction probability distribution before and after instilling the target knowledge, employing metrics such as entropy and KL-divergence. 
Introducing our metrics, we first assess their accuracy in comparison to previous ranking-based methods, surpassing them by over $35\%$ in a synthetic experiment. 
Then, we explore two prominent methods of knowledge instillation, discovering that LLMs exhibit limitations in capturing new knowledge under specific circumstances for one of these methods. 
Lastly, we demonstrate the applicability of our methods in extracting unlearned and mislearned facts in LLMs through their application to in-context learning. 
We make code and data for all methods and experiments in this paper publicly available.\footnote{\url{https://github.com/rit-git/lm-know}}
\end{abstract}

\section{Introduction}
Large language models (LLMs) have demonstrated significant success in various downstream tasks \citep{devlin2019bert, brown2020language}. 
These models are trained on massive amounts of text, and they encode world knowledge in their parameters, making it possible to solve downstream tasks effectively. 
Therefore, it is essential to comprehend and quantify the extent of LLMs' knowledge about various facts. 

\begin{figure*}[t]
    \centering
        \captionsetup[subfigure]{justification=centering}
        \begin{subfigure}{.84\linewidth}
            \centering
            \includegraphics[width=\textwidth]{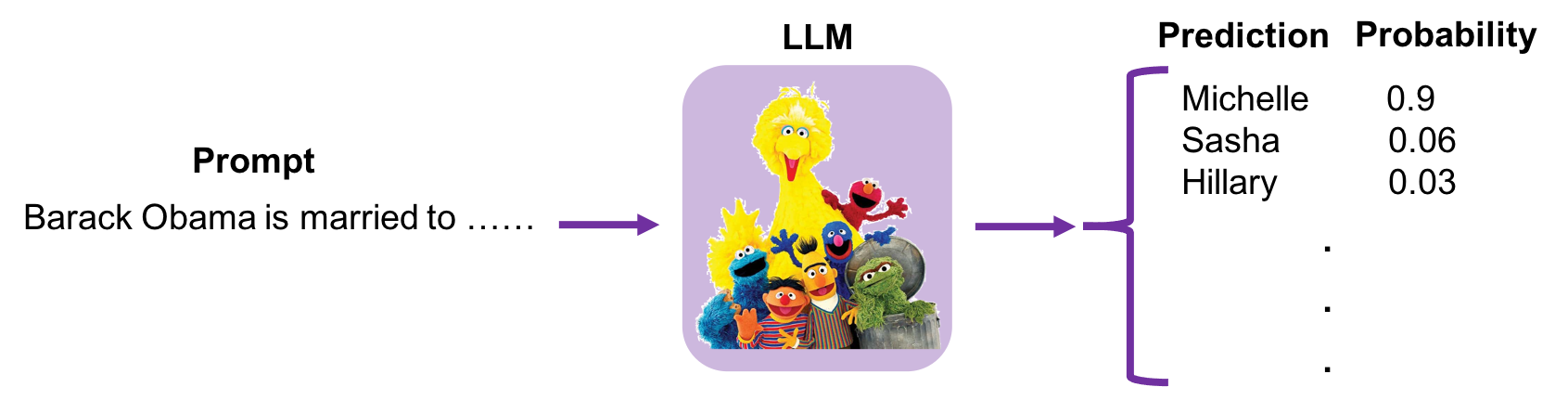}
            \caption{Probing.}
            \label{fig:prob}
            \vspace{0.2cm}
        \end{subfigure}
        \begin{subfigure}{.47\linewidth}
            \centering
            \includegraphics[width=\textwidth]{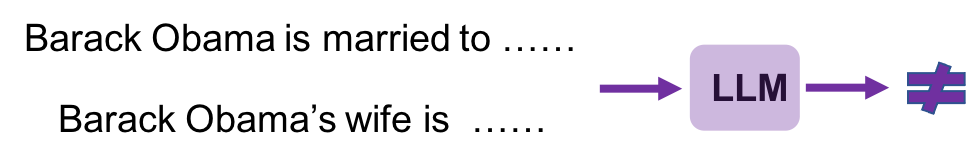}
            \vspace{0.45cm}
            \caption{Sensitivity.}
            \label{fig:sen}
        \end{subfigure}
        \hspace{0.5cm}
        \begin{subfigure}{.47\linewidth}
            \centering
            \includegraphics[width=\textwidth]{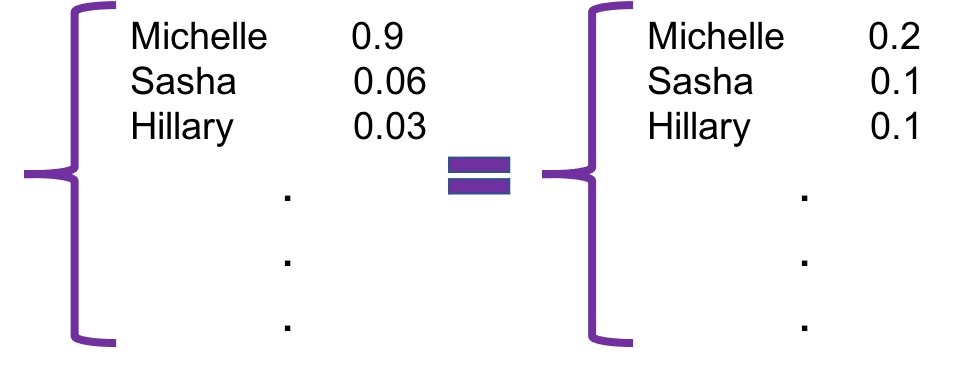}
            \caption{Incapability in capturing knowledge.}
            \label{fig:incapability}
        \end{subfigure}
    \caption{\textbf{Using ranking-based metrics in probing and its current limitations in measuring factual knowledge}: The current measurement methods predominantly rely on ranking metrics, which are vulnerable to prompt perturbation, lack the ability to effectively capture knowledge, and are mostly limited to binary representations of knowledge.}
    \label{fig:overview}
    \minipostspace
\end{figure*}

Over the last few years, probing techniques have been introduced to assess the knowledge of LLMs \citep{petroni2019language, vulic2020probing,cao2021knowledgeable, alkhamissi2022review}. 
These techniques mostly defined as fill-in-the-blank tasks that measure the model's knowledge by ranking its predictions (overview of probing is depicted in Figure~\ref{fig:prob}). 
However, while these approaches provide a useful binary representation of knowledge by incorporating ranking metrics, there are several fundamental issues with this procedure. 
Firstly, knowledge is not binary and cannot be fully captured by such a representation. 
Secondly, ranking metrics are often highly sensitive to the specific prompts used, leading to potential biases in the assessment (an example is provided in Figure~\ref{fig:sen}). 
Finally, these metrics may not be able to capture knowledge accurately; as highlighted in Figure~\ref{fig:incapability}, the gold label ranking is the same for these two distributions, despite the fact that these two predictions exhibit a complete different level of knowledge regarding the target fact. 
Therefore, to gain a more comprehensive understanding of LLMs' knowledge, it is necessary to develop better metrics that go beyond the binary notion of knowledge and account for these limitations.

In this work, we propose a new framework that utilize measurements of knowledge derived from information theory. 
By examining the probability distribution of language models' vocabulary when predicting a blank, we first introduce the concept of prompt's uncertainty. 
Then, using the intuition that a LLM know a fact if the prompt's uncertainty remains the same after instilling that fact into the model, we introduce our measurements by incorporating information theory based metrics such as entropy and KL-divergence, capturing uncertainty, to measure knowledge in language models.

To instill a knowledge into a language model, we examine two approaches: (1) Explicit instillation, by directly including the fact in the prompt used in the probing, and (2) Implicit instillation, by fine-tuning the model on that specific fact.  
One important research question that we aim to address here is: when it is appropriate to instill a  knowledge explicitly. 
This is a particularly critical question because the implicit method of instilling information can be very costly and may not be feasible for certain in-context learning-based models such as GPT-3 \citep{brown2020language} and GPT-4 \citep{openai2023gpt-4}.

To demonstrate validity and applicability of our proposed knowledge metrics, we conduct various experiments. 
We first investigate the accuracy of our metric, as compared to ranking methods surpassing them by around $30\%$ in a synthetic experiment. 
We then examine the differences between the two methods of knowledge instillation and determine situations where the explicit method of instillation is inadequate. 
Additionally, we explore the potential applications of our methods for in-context-learning based models through two distinct tasks: (1) factual alignment, where we address the question of whether it is necessary to explicitly provide a certain fact in the prompt for it to appear in the generated text. 
And, (2) avoiding hallucination, by calculating the correlation between the LLM's knowledge and the occurrence of hallucinated versus accurately generated facts.

\section{Measuring Factual Knowledge}
Over the past few years, probing has emerged as a fundamental approach for assessing the factual knowledge encoded in the parameters of large language models.
The aim of probing is to present different snippets of factual information (prompts) to LLMs, asking them to predict a masked/blank section of the information. 
An example of this process is depicted in Figure~\ref{fig:overview}, where we probe the LLM for information about Barack Obama's marriage partner. 
By extracting the model's probability distribution over the vocabulary for the masked/blank token, prior research has employed various ranking metrics to determine the level of the large language models' understanding for a given factual knowledge.

Despite the valuable insights provided by probing in understanding LLMs' knowledge, it is subject to several limitations. These limitations include: 
(1) heavy sensitivity to the specific wording of each prompt, which can impact the model's performance.
(2) Binary representation of knowledge does not align with real-world knowledge representation.
And (3) the exclusive use of gold label ranking, while disregarding the predicted probability of every other words in the vocabulary, results in the inability to differentiate between various levels of a model knowledge about different facts. These limitations are illustrated in Figure~\ref{fig:overview}.

In this work, to overcome these limitations, we adopt knowledge measurements from information theory. 
More specifically, we use entropy to define the prompt's uncertainty as:
\begin{align}
    H(\text{prompt}) = - \sum_{k \in V} P_k log(P_k)
\end{align}
Where $V$ represents the vocabulary. Using the entropy of a prompt, we can intuitively determine whether the LLM knows the fact in question ($f$), if:
\begin{align}
    H(\text{prompt}) \sim H(\text{prompt}|\text{instilling $f$ into LLM})
\end{align}
Using this intuition, we can measure the knowledge contained in the LLM about $f$ by computing $H(\text{prompt}) - H(\text{prompt}|\text{instilling $f$ into LLM})$. 
By employing this metric to measure knowledge, we can move beyond binary representation and instead capture a more nuanced understanding of knowledge. This allows us to mitigate the sensitivity to prompts by accounting for the impact of subtracting uncertainty terms on the robustness. Additionally, since entropy considers the complete probability distribution, we can assess knowledge more comprehensively rather than relying solely on gold label rankings.

\begin{figure}[t]
    \centering
    \includegraphics[width=0.95\columnwidth]{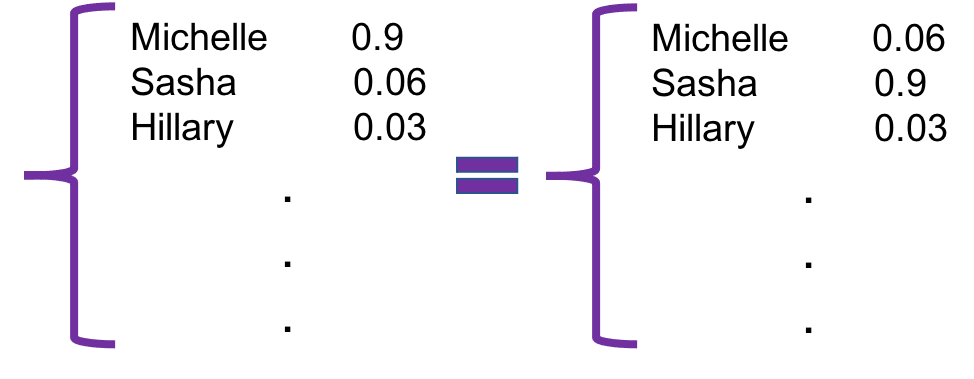}
    \caption{Entropy based metrics cannot capture order in the probability distribution.}
    \label{fig:ent-issue}
\end{figure}

While our entropy-based metric can address the limitations of ranking metrics, it still has certain limitations.
For instance, entropy cannot account for the order in the predicted probabilities of words in the vocabulary, which may hinder accurate knowledge measurement (an example of this issue is provided in Figure~\ref{fig:ent-issue}). 
To overcome this limitation, we explore the application of KL-divergence for measuring the LLM's knowledge of the target fact ($f$). 
KL-divergence is defined as $KL(P||Q)$, where $P$ represents the initial probability distribution of the prompt over the vocabulary, and $Q$ represents the probability distribution of the prompt after instilling $f$ into the LLM. 
More specifically, we will have:
\begin{align}
    KL_{\text{score}}(\text{prompt}) = - \sum_{k \in V } P_k log(\frac{P_k}{Q_k})
\end{align}

To approximate the predicted probability distribution of language models such as GPT-3 when the full vocabulary ($V$) is not accessible, we adopt a specific approach. 
First, we obtain the top-k probable tokens (with their predicted probability) from the model before knowledge instillation ($V_b$) and after knowledge instillation ($V_a$). 
Then, we approximate the vocabulary by creating a new vocabulary ($V'$) that includes only the tokens present in $V_a$ and $V_b$, along with an out-of-vocabulary (oov) token. 
The size of $V'$ is determined by the union of $V_a$ and $V_b$, denoted as $|V_a \cup V_b| + 1$. 

Next, we uniformly distribute the missing probability mass from the sum of the top-k predictions among the remaining tokens in $V'$ (for both, before and after knowledge instillation). 
This ensures that the probability distribution remains consistent even when some tokens are missing from the $V_a$ and $V_b$. 
Finally, we utilize this resultant distribution for our factual knowledge measurements. This approach allows us to approximate the predicted probability distribution of the language model despite not having access to the full vocabulary.

\section{Implicit vs Explicit Knowledge Instillation}
We consider two form of knowledge instillation for LLMs:

\paragraph{Explicit} knowledge instillation refers to incorporating knowledge into an LLM by explicitly including it in the prompt. 
For instance, to incorporate information about Barack Obama's marriage into an LLM, instead of asking ``Barack Obama is married to ...'', we would prompt the LLM by probing ``Barack Obama is married to Michelle Obama. Barack Obama is married to ...''.

\paragraph{Implicit} knowledge instillation involves incorporating knowledge into an LLM by fine-tuning the LLM on that particular knowledge. 
For example, we can implicitly incorporate information about Barack Obama's marriage into BERT by fine-tuning it on the prompt ``Barack Obama is married to [MASK]''.

Our goal in this work is to answer the research question of when it is appropriate to instill information explicitly as opposed to through fine-tuning. 
This is an important question to address as fine-tuning (implicit instillation) can be costly and may not even be feasible for LLMs such as GPT-3 \citep{brown2020language} and GPT-4 \citep{openai2023gpt-4}. 
By comparing the two forms of knowledge instillation, we aim to determine the conditions under which explicit instillation is accurate and effective.

\section{Experiment Setup}

\paragraph{Datasets}
We conducted various experiments on fact-checking benchmarks T-REx \citep{elsahar2018t} and LAMA \citep{petroni2019language} to assess different knowledge metrics. To compare implicit and explicit knowledge instillation in Section \ref{sec:impvsexp}, we randomly sampled 100 facts from T-REx for each relations appeared in LAMA (more details in Appendix).

\paragraph{Models}
For our evaluations, we utilized two popular large language models, BERT \citep{devlin2019bert} and T5 \citep{raffel2019exploring}, to gauge the accuracy of various knowledge metrics and to compare the effectiveness of explicit and implicit knowledge instillation techniques. Additionally, we employed InstructGPT (text-davinci-003) \citep{ouyang2022training} and FLAN-T5 (XL) \citep{chung2022scaling} to investigate the applicability of our proposed methods across different tasks for in-context learning based models.

\section{Experiments}
In this section, we first evaluate the accuracy of various metrics in measuring an LLM's knowledge of a fact. 
We then explore the differences between implicit and explicit knowledge instillation, determining when it is appropriate to instill a knowledge explicitly. 
Lastly, we examine the application of our proposed metrics in in-context learning based models.

\begin{table}
\small
\centering
\begin{tabular}{lrr}
\toprule 
\bf Metrics& \bf BERT& \bf T5\\
\midrule
Ranking & 51.6& 30.9 \\
Entropy& 72.2& 66.4\\
KL-Divergance&\bf 74.5& \bf 67.8\\
\bottomrule
\end{tabular}
\caption{Accuracy of knowledge metrics in correctly assigning higher level of knowledge to the facts with better representation is evaluated by fine-tuning LLMs on the synthetically provided facts.}
\label{tab:acc}
\end{table}

\begin{figure*}[t!]
\captionsetup[subfigure]{justification=centering}
\begin{subfigure}{.52\linewidth}
    \centering
    \includegraphics[width=\textwidth]{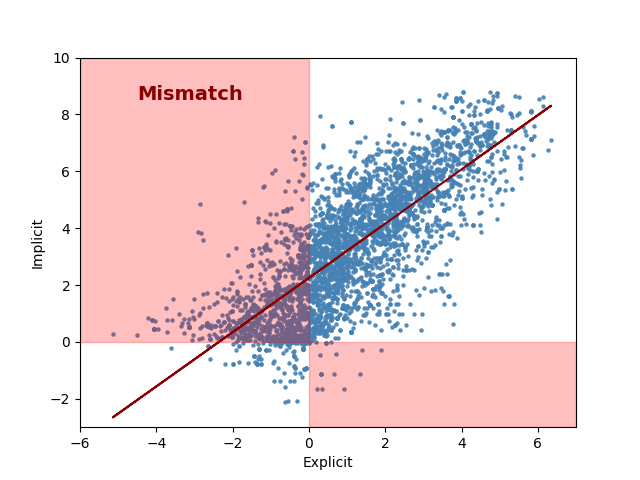}
    \caption{Entropy.}
\end{subfigure}
\begin{subfigure}{.52\linewidth}
    \centering
    \includegraphics[width=\textwidth]{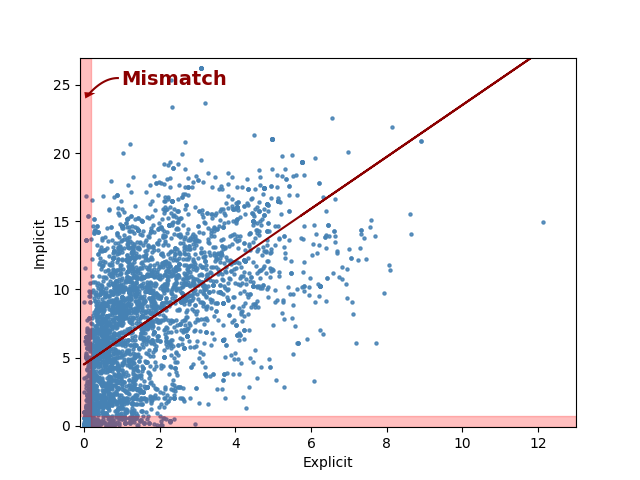}
    \caption{KL-Divergance.}
\end{subfigure}
    \caption{The correlation between explicit and implicit knowledge instillation using entropy and KL-divergence metric for \textit{BERT} language model. For the entropy metric, mismatch happens when the sign of the metric differs between implicit and explicit instillation. In the KL-divergence case, the mismatch arises when the metric for implicit instillation is significantly higher or lower than that of explicit instillation.}
    \label{fig:bert-scat}
\end{figure*}
\subsection{Accuracy of Knowledge Measurements} 
As we lack access to the amount of knowledge that language models possess for any given fact, we conducted a synthetic experiment to evaluate the accuracy of different metrics.  
We fine-tuned BERT/T5 on a filling-the-blank task to create a gold label for these models' knowledge on a specific fact.  
For each relation in LAMA, we collected instances where ranking metrics performed poorly, i.e., facts that the models lacked knowledge of. 
Then, we iteratively removed parts of the prompts corresponding to those facts for each relation to create instances that conveyed less and less information to the model. 
For instance, for the relation \textit{is married to}, starting with instances (1) John is married to [Niki], (2) Mark is married to [Emma], (3) Liam is married to [Ava], (4) William is married to [Sophia], and (5) Noah is married to [Katherine], we modified them to (1) John is married to [Niki], (2) Mark married to [Emma], (3) Liam to [Ava], (4) William [Sophia], and (5) Noah. 
We then fine-tuned the models to predict the object over the modified instances. 
Finally, we evaluated the fine-tuned models over the initial examples and calculated the average pairwise accuracy of metrics in selecting the instance that the model should have more knowledge about (e.g., the model's knowledge of fact (1) should be higher than that of fact (2)).
Table \ref{tab:acc} presents the accuracy of different knowledge metrics. 
The results reveal that KL-divergence and entropy-based metrics surpass ranking methods by more than $20\%$ and $35\%$ respectively in BERT and T5, showcasing the superior accuracy of our proposed metrics in capturing the factual knowledge of LLMs.  Additionally, KL-divergence exhibits a slight advantage over entropy in both LLMs.

\subsection{Implicit vs Explicit Knowledge Instillation}
\label{sec:impvsexp}
\begin{figure*}[t!]
\captionsetup[subfigure]{justification=centering}
\begin{subfigure}{.52\linewidth}
    \centering
    \includegraphics[width=\textwidth]{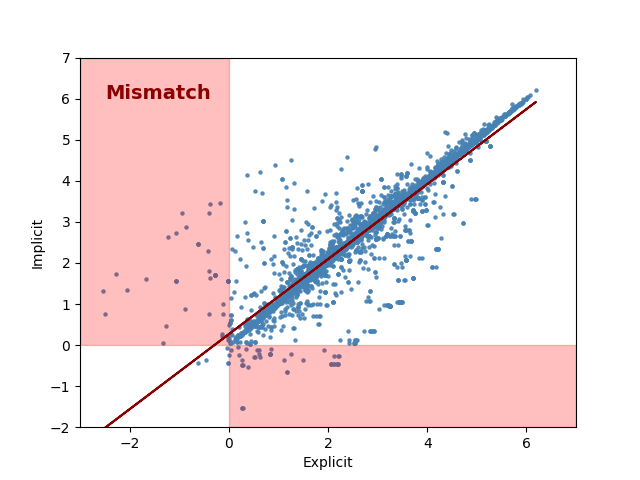}
    \caption{Entropy.}
\end{subfigure}
\begin{subfigure}{.52\linewidth}
    \centering
    \includegraphics[width=\textwidth]{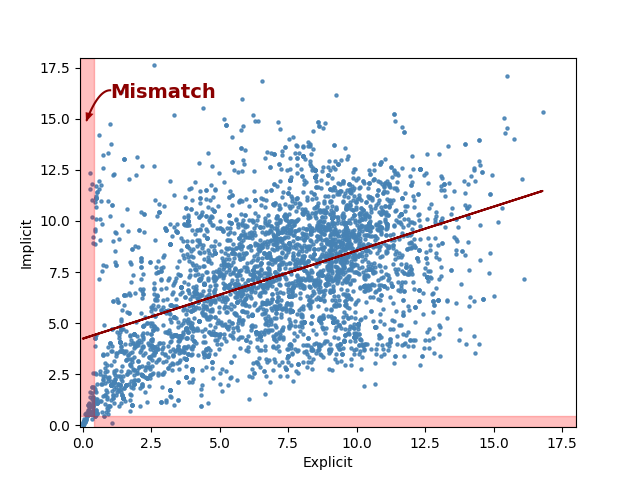}
    \caption{KL-Divergance.}
\end{subfigure}
    \caption{The correlation between explicit and implicit knowledge instillation using entropy and KL-divergence metric for \textit{T5} language model. The mismatch regions are being identified as before.}
    \label{fig:T5-scat}
\end{figure*}

Given the high cost and sometimes infeasibility of implicit knowledge instillation (fine-tuning), it is important to determine when explicit knowledge instillation can be a viable alternative. 
To address this question, we conducted a comparison between two different knowledge instillation methods using BERT and T5 language model using our proposed metrics over the LAMA benchmark. 
The resulting scatter plots, depicting instances of implicit versus explicit instillation for BERT and T5, are shown in Figures \ref{fig:bert-scat} and \ref{fig:T5-scat}, respectively. 
Based on these results, we first observe a strong correlation between implicit and explicit knowledge instillation. 
Furthermore, T5 exhibits a higher/better level of correlation between these two methods compared to BERT, indicating that we can estimate implicit knowledge with greater accuracy in this model using explicit instillation. 

Notably, there are specific regions in the plot where a mismatch occurs between the two forms of instillation. 
Specifically, for the entropy metric, these regions correspond to instances where the sign of the metrics differs between implicit and explicit instillation. 
In the case of KL-divergence, the mismatch arises when the metric for implicit instillation is significantly higher or lower than that of explicit instillation. 
Upon further investigation of the instances falling into these mismatched areas, we find that the majority of them are samples with labels related to location (e.g., {\relfont has capital} relation) or language (e.g., {\relfont has official language} relation) for both BERT and T5. 
This demonstrates that we cannot approximate implicit instillation with the explicit approach for these types of relations. 
Additionally, T5 exhibits fewer instances of mismatch compared to BERT. 
Considering that both BERT and T5 share relation types that require implicit instillation for accurately inheriting knowledge, the question that remains is whether these problematic relations also affect in-context learning models like InstructGPT. 
We address this question in the next section.

\subsection{In-Context Learning Based Applications}
While our proposed methods have demonstrated superior performance compared to ranking-based metrics, it remains unclear whether they have practical utility beyond the realm of analyzing LLMs knowledge. 
This is especially true for in-context learning based methods, for which implicit instillation may not be a viable option. 
Therefore, in this section, we aim to explore the real-world applications of our metrics in two tasks: (1) factual alignment, where we investigate how our metrics can ensure that specific facts appear in LLMs' generation, and (2) avoiding hallucination, by measuring the correlation between our knowledge metrics for hallucinated and non-hallucinated facts.

\paragraph{Factual Alignment}
Factual alignment refers to the task of ensuring that a specific fact appears in the generated output of an LLM. 
This task is particularly important in cases where a more accurate or controllable generation is required. 
To investigate factual alignment using our knowledge metrics, we ask the LLM to write a summary about an entity and categorize the facts about that entity into two categories: (1) facts that appear in the summary, and (2) facts that didn't appear in the generated output. 

To conduct the factual alignment experiment, we selected a set of the most popular human and non-human entities and their corresponding facts from the T-REx benchmark. 
We gather 500 entities and their 5175 corresponding facts from T-REx benchmark. 
We prompted the LLMs to generate a paragraph about each entity by using the prompt, "Write a paragraph about [entity]." 

\paragraph{Hallucination} which refers to the incorrect, nonsensical, or incoherent facts in the generated text, can hinder the real-world adoption of LLMs in various applications. 
In here, our objective is to investigate whether it is feasible to use our metrics to identify the facts that are probable to be hallucinated by the LLM. 
Our conjecture is that the hallucinated facts are typically those that the model has less information about, which is what we investigate here. 
We utilize entities, their associated facts, and the generated paragraphs obtained in \textit{factual alignment} experiments to examine the effectiveness of our metrics in accurately detecting fabricated facts.

Before investigating the applicability of our metrics in factual alignment and detecting hallucination, we need to define a model that can predict whether a given fact \textit{appeared}, \textit{didn't appear}, or \textit{appeared incorrectly} (hallucinated) in a given paragraph. 
To accomplish this, we fine-tune a RoBERTa-based \citep{liu2019roberta} classifier by extracting facts from LAMA \citep{petroni2019language} and their corresponding prompts from T-REx dataset \citep{elsahar2018t}. 
T-REx provides prompts in the form of paragraphs, where the facts can appear explicitly or implicitly within these paragraphs. 
In order to gather data for the \textit{didn't appear} class, we replace the object of the fact by randomly sampling from all the objects connected to the subject of our target fact. 
Similarly, for the \textit{appeared incorrectly} class, we replace the object of the fact by randomly sampling from all the objects that appear in the graph with that relation. 
Our training, development, and test sets consist of 5000, 1000, and 1000 samples, respectively. 
The discriminator achieves $90.4\%$ accuracy on the test data.

To enhance the accuracy of our discriminator when applied to generated text, we only consider predictions with a confidence level exceeding $0.95$. 
Additionally, we evaluate the accuracy of the discriminator on generated text in a user study by randomly selecting 100 instances for each class and model and asking three participants to classify the given fact and generated paragraph pairs. 
We then employ majority voting to determine the classification for each pair. 
The result of the user study is presented in Table \ref{tab:user}, demonstrating that our discriminator achieves over $81\%$ accuracy for all classes in both InstructGPT and Flan-T5 LLMs.

\begin{table}
\small
\centering
\begin{tabular}{lrr}
\toprule 
&IntructGPT&FLAN-T5\\
\midrule
Appeared&100&92\\
Didn't Appear&86&95\\
Hallucinated&82&81\\
\bottomrule
\end{tabular}
\caption{The accuracy of the discriminator in classifying facts according to their appearance in generated paragraphs is evaluated through a user study.}
\label{tab:user}
\end{table}

\begin{table*}[!ht]
\small
\centering
\begin{tabular}{lrrrrrr}
\toprule 
\multirow{2}{*}{\bf Relations} & \multicolumn{3}{c}{\bf IntructGPT}&  \multicolumn{3}{c}{\bf FLAN-T5}\\
\cmidrule(lr){2-4}
\cmidrule(lr){5-7}
&Appeared&Didn't Appear&Hallucinated&Appeared&Didn't Appear&Hallucinated\\
\midrule
shares border with & \gray 0.252 &\gray 0.155 &\gray0.162 & \gray 0.725 & \gray 1.147 & \gray 0.64 \\
 official language &\gray 1.737 & \gray 2.823 & \gray 2.407 & \green 9.327 & \green 6.787 & \green - \\
 named after &\green 0.056 & \green0.384 & \green0.158 & \gray 12.109 &\gray 11.232 &\gray 7.941 \\
 part of &\green 0.001 &\green 0.0 &\green 0.017 & \gray 10.951 &\gray 9.13 &\gray 13.083 \\
 capital & \gray 1.736 & \gray 2.898 & \gray 1.68 & \green 3.375 & \green 6.33 & \green 9.599 \\
 diplomatic relation &\green 0.035 &\green 0.133 &\green 0.339 &\gray 3.215 &\gray 1.956 &\gray 3.45 \\
 sister city & \green- &\green 5.196 &\green 1.621 &\green - &\green 9.903 &\green - \\
 continent & \green0.175 &\green 0.002 &\green 0.078 & \gray 7.363 &\gray 5.378 &\gray 5.938 \\
 capital of & \gray 1.242 & \gray 0.72 &\gray 0.793 & \gray 8.504 &\gray 8.275 &\gray 7.207 \\
 place of birth & \gray 1.335 &\gray 1.681 &\gray 2.501 &\gray - &\gray 9.144 &\gray 7.618 \\
 genre &\green 0.025 &\green 0.715 &\green 0.028 &\green - &\green - &\green 3.862 \\
 located in the admin territory&\green 0.147 &\green - & \green0.005 &\gray 4.862 &\gray 4.945 &\gray 6.233 \\
 country &\green 0.003 &\green - &\green 0.007 &\green 2.84 &\green 5.93 &\green 1.739 \\
 has part &\green - & \green- &\green 0.004 &\green - &\green - &\green 10.635 \\
 religion &\green - &\green - &\green 5.938 &\gray - &\gray - &\gray - \\
 country of citizenship &\green 1.999 &\green - &\green 0.584 &\gray 1.542 &\gray - &\gray 2.631 \\
 field of work &\gray 0.333 &\gray - &\gray 0.309 & \green3.364 &\green - &\green 6.093 \\
 occupation &\green 0.119 & \green - &\green 0.367 &\green - &\green - &\green 5.662 \\
 position held &\gray 0.938 &\gray - &\gray 0.91 &\green 2.434 &\green - &\green 8.29 \\
 work location &\green 0.116 &\green - &\green 0.355 & \green4.94 &\green 9.411 &\green 3.687 \\
 instrument &\gray 0.017 &\gray - &\gray 0.012 &\green - &\green - &\green 7.387 \\
 place of death &\green 0.461 &\green - &\green 0.135 & \gray 0.881 &\gray 0.912 &\gray 2.09 \\
 position played &\green 1.41 &\green - &\green 0.136 & \green - &\green - &\green 6.054 \\
 headquarters location &\green 0.564 &\green - &\green - &\green 6.692 &\green - &\green - \\
 location of formation &\green 0.827 &\green - &\green - &\gray - &\gray - &\gray - \\
 employer &\green 0.004 \green&\green - &\green - & \gray 2.212 &\gray - &\gray 1.855 \\
 member of &\green 0.056 &\green - &\green - &\green - &\green - &\green 7.075 \\
 instance of &\gray - &\gray - &\gray - &\green - &\green 0.899 &\green - \\
 developer &\gray - &\gray - &\gray - &\green - &\green 6.875 &\green - \\
 language of work or name &\gray - &\gray - &\gray - &\green - &\green - &\green 12.251 \\
 country of origin &\gray - &\gray - &\gray - &\green 1.838 &\green - &\green 10.112 \\
 original language of work &\gray - &\gray - &\gray - & \green 0.489 &\green - &\green 13.142 \\
 owned by &\gray - &\gray - &\gray - &\green 0.165 &\green - &\green - \\
\bottomrule
\end{tabular}
\caption{\textbf{Per-relation breakdown} of three classes of facts,  categorized by their appearance in the generated paragraphs produced by InstructGPT and FLAN-T5, is presented to evaluate the effectiveness of the \textit{KL-divergence} metric in distinguishing between facts across these classes (bigger numbers indicate the lower amount of knowledge). Relations in which our metric demonstrates effective differentiation between different fact classes are highlighted in green.}
\label{tab:factual-kl}
\end{table*}

The results of the factual alignment and Hallucination experiments can be found in Table \ref{tab:factual-kl} (results of measuring knowledge based on entropy is provided in Appendix). 
The objective of this analysis is to identify relation types that our metrics can potentially differentiate among the three classes: \textit{appeared}, \textit{didn't appear}, and \textit{hallucinated} (appeared incorrectly) facts. 
In the table, we have highlighted relation types where there is a meaningful difference in our metrics across these classes in green. 
Firstly, it is evident that most of the relations where our metrics are unable to differentiate between the classes involve location or language as their object. 
Additionally, when comparing \textit{appeared} with \textit{hallucinated} facts, we observe that for relations with a location as the object, the model possesses more knowledge about \textit{hallucinated} facts in comparison to \textit{appeared} ones. 
Conversely, for other types of relations, the model demonstrates a higher knowledge level for \textit{appeared} facts. 
Moreover, except for a few relations involving location and language as their object, the LLMs exhibit significantly lower knowledge for \textit{didn't appear} facts when compared to the other two classes.

Further analysis of the results reveals interesting observation in relation to the need for factual alignment and probability of hallucination for InstructGPT and FLAN-T5. The relations {\relfont admin territory, country, country of citizenship, occupation, work location, place of death, position played, headquarters location, location of formation, employer, member of} show a lower requirement for explicit knowledge instillation to appear in the generated output in InstructGPT. On the other hand, for FLAN-T5, the relations {\relfont field of work, position held, headquarters location, country of origin, original language of work, owned by} exhibit a similar characteristic.
Moreover, certain relations demonstrate higher resistance to hallucination in InstructGPT and FLAN-T5. Specifically, the relations {\relfont headquarters location, location of formation, employer, member of} exhibit a greater resistance to hallucination in InstructGPT, while the relations {\relfont official language, sister city, headquarters location, instance of, developer, owned by} demonstrate a higher resistance to hallucination in FLAN-T5.

Lastly, upon examining samples where the injected information did not result in accurate predictions, we can identify relations where explicit instillation alone is insufficient. 
Since we cannot fine-tune these models (at least for InstaructGPT) and compare implicit with explicit directly, we consider failure in explicit knowledge instillation in cases where the label does not appear within the top 5 predicted outputs. 
Similar to the previous analysis, approximately $80\%$ of the mispredicted samples even after explicit instillation were associated with location or language queries for both InstructGPT and FLAN-T5. Moreover, these relations primarily consist of the ones highlighted in red in Table \ref{tab:factual-kl}.

\section{Related Works}
Large language models (LLMs) have emerged as the central focus of recent advancements and applications in NLP. 
Given their extensive repository of factual knowledge, effectively harnessing these models for downstream tasks necessitates accurate measurements of their inherited knowledge.

\paragraph{Measuring factual knowledge in LLMs} 
The significance of ensuring factual correctness in LLMs has received considerable attention due to its critical role in determining the applicability of language models. 
Previous studies \citep{petroni2019language,alkhamissi2022review} have explored the quantification of factual knowledge in LLMs by assessing their understanding of facts in knowledge bases using ranking metrics. 
In a different approach, \citet{varshney2022investigating} incorporate question answering as a means to measure the uncertainty of LLMs regarding specific facts.  
Furthermore, recent works \citep{kadavath2022language, lin2022teaching} have adopted self-evaluation techniques by querying LLMs themselves to assess their certainty regarding factual knowledge.

\paragraph{Factual knowledge in in-context learning} 
Given the remarkable success of in-context learning based LLMs \citep{brown2020language,chowdhery2022palm, touvron2023llama} across various NLP applications, factual knowledge serves as an invaluable resource for evaluating and guiding the generated output of these models. 
\citet{cohen2023crawling} employed prompting to crawl internal knowledge within these models, thereby constructing knowledge bases. 
Authors in \citet{peng2023check} augmented factual knowledge into LLMs to enhance the accuracy of their output. 
Furthermore, recent studies \citep{goodrich2019assessing, shuster2021retrieval,ji2023survey} utilize factual knowledge to detect and mitigate hallucination in the generated output of LLMs.

\section{Conclusion}
In this paper, we introduced novel metrics for measuring factual knowledge in large language models (LLMs) compensating for the shortcomings of existing ranking-based methods. Our results revealed that our proposed metrics outperformed traditional ranking-based approaches, providing more accurate assessments of factual knowledge in LLMs.  
Additionally, we explored the distinction between implicit and explicit knowledge instillation in LLMs. Through comprehensive experiments, we observed cases where explicit knowledge instillation alone was inadequate, highlighting the need for fine-tuning. 
These cases primarily revolve around location and language-related queries, emphasizing the intricate nature of these types of facts and the challenges they pose for explicit instillation. 
This finding contributes to our understanding of the interplay between implicit and explicit knowledge in LLMs. 

We investigated the application of our metrics in two crucial areas: factual alignment and hallucination detection for in-context learning based models. 
Upon applying our proposed metrics to these tasks, we exhibit promising results, offering valuable insights into aligning generated output with factual knowledge and identifying and mitigating hallucinated facts. 
Furthermore, our observations indicate that even in these significantly enhanced LLMs, explicit knowledge instillation continues to encounter challenges when it comes to location and language-related queries. 
All code and data necessary to reproduce the results reported in this paper is available at: \url{https://github.com/rit-git/lm-know}.

\section*{Acknowledgements}
We would like to thank Tom Mitchell, Estevam Hruschka, Nikita Bhutani, Eser Kandogan, and Yasaman Razeghi for their valuable comments. 

\clearpage
\bibliography{emnlp2023}
\bibliographystyle{acl_natbib}


\appendix

\section{Experimental Details}
\paragraph{Datasets}
Our experimental evaluations involved fact-checking benchmarks such as T-REx \citep{elsahar2018t}, which is a curated subset of Wikipedia triples aligned with corresponding Wikipedia abstracts. T-REx encompasses a vast collection of 11 million triples and 3.09 million Wikipedia abstracts, covering over 600 distinct Wikidata predicates. 
To facilitate the mapping of triples from T-REx to natural language expressions, we employed the LAMA framework introduced by \citet{petroni2019language}. LAMA provides natural language templates specifically designed for 41 predicates derived from the T-REx benchmark.

\paragraph{Models}
For our evaluations, we utilized two popular models, BERT \citep{devlin2019bert} and T5 \citep{raffel2019exploring}, to gauge the accuracy of various knowledge metrics and to compare the effectiveness of explicit and implicit knowledge instillation techniques. Additionally, we employed InstructGPT (text-davinci-003) \citep{ouyang2022training} and FLAN-T5 (XL) \citep{chung2022scaling} to investigate the applicability of our proposed methods across different tasks using in-context learning-based models.

\section{In-Context Learning Results Based on Entropy Metric}
The results of the factual alignment and Hallucination experiments for entropy based metric can be found in Table \ref{tab:factual-ent}. 
In the table, we have highlighted relation types where there is a meaningful difference in our metrics across these classes in green. 
Firstly, it is evident that most of the relations where our metrics are unable to differentiate between the classes involve location or language as their object. 
Additionally, when comparing \textit{appeared} with \textit{hallucinated} facts, we observe that for relations with a location as the object, the model mostly possesses more knowledge about \textit{appeared} facts in comparison to \textit{appeared} ones.

Further analysis of the results reveals interesting trends in relation to the need for factual alignment and hallucination for InstructGPT and FLAN-T5. The relations {\relfont admin territory, country, field of work, work location, instrument, headquarters location, location of formation, employer, member of} show a lower requirement for explicit knowledge instillation to appear in the generated output in InstructGPT. On the other hand, for FLAN-T5, the relations {\relfont headquarters location, employer, country of origin, original language of work, owned by} exhibit a similar characteristic.
Moreover, certain relations demonstrate higher resistance to hallucination in InstructGPT and FLAN-T5. Specifically, the relations {\relfont headquarters location, location of formation, employer, member of} exhibit a greater resistance to hallucination in InstructGPT, while the relations {\relfont official language, sister city, headquarters location, instance of, developer, owned by} demonstrate a higher resistance to hallucination in FLAN-T5.

\begin{table*}[!ht]
\small
\centering
\begin{tabular}{lrrrrrr}
\toprule 
\multirow{2}{*}{\bf Relations} & \multicolumn{3}{c}{\bf IntructGPT}&  \multicolumn{3}{c}{\bf FLAN-T5}\\
\cmidrule(lr){2-4}
\cmidrule(lr){5-7}
&Appeared&Didn't Appear&Hallucinated&Appeared&Didn't Appear&Hallucinated\\
\midrule
 shares border with & \red 0.164 &\red 0.127 &\red 0.111 &\red 1.245 &\red 0.929 &\red 0.948 \\
 official language &\red 0.318 &\red 0.372 &\red 0.427 & \green 1.221 & \green 0.835 &\green - \\
 named after &\green 0.071 &\green 0.272 &\green 0.141 &\green 2.441 &\green 1.831 &\green 1.08 \\
 part of &\green 0.01 &\green 0.006 &\green 0.076 &\red 2.417 &\red 2.416 &\red 2.372 \\
 capital &\red 0.202 &\red 0.22 &\red 0.305 & \green 0.408 &\green 1.155 &\green 0.746 \\
 diplomatic relation &\red 0.111 &\red 0.189 &\red 0.204 &\red 0.665 &\red 0.518 &\red 0.803 \\
 sister city &\red - &\red 0.67 &\red 0.48 &\green - &\green 0.511 &\green - \\
 continent &\green 0.099 &\green 0.003 &\green 0.122 &\red 1.61 &\red 1.487 &\red 1.578 \\
 capital of &\red 0.217 &\red 0.565 &\red 0.191 &\green 1.822 &\green 2.176 &\green 0.905 \\
 place of birth &\red 0.192 &\red 0.392 &\red 0.346 &\green - &\green 0.872 &\green 1.146 \\
 genre &\green 0.088 &\green 0.713 &\green 0.1 &\green - &\green - &\green 1.459 \\
 located in the admin territory &\green 0.137 &\green - &\green 0.014 &\red 1.621 &\red 1.907 &\red 1.027 \\
 country &\green 0.025 &\green - &\green 0.039 &\green 2.393 &\green 0.762 &\green 1.357 \\
 has part &\green - &\green - &\green 0.034 &\green - &\green - &\green 1.6 \\
 religion &\green - &\green - &\green 0.466 &\red - &\red - &\red - \\
 country of citizenship &\red 0.336 &\red - &\red 0.429 &\red 1.104 &\red - &\red 0.859 \\
 field of work &\green 0.267 &\green - &\green 0.634 &\red 1.476 &\red - &\red 1.144 \\
 occupation &\red 0.246 &\red - &\red 0.273 &\green - &\green - &\green 1.224 \\
 position held &\red 0.354 &\red - &\red 0.336 &\red 1.674 &\red - &\red 1.241 \\
 work location &\green 0.131 &\green - &\green 0.221 &\green 1.78 &\green 0.539 &\green 2.736 \\
 instrument &\green 0.046 &\green - &\green 0.017 &\green - &\green - &\green 1.34 \\
 place of death &\red 0.206 &\red - &\red 0.159 &\red 1.305 &\red 1.289 &\red 1.297 \\
 position played&\red 0.271 &\red - &\red 0.399 &\green - &\green - &\green 0.525 \\
 headquarters location &\green 0.498 &\green - &\green - &\green 1.387 &\green - &\green - \\
 location of formation &\green 0.288 &\green - &\green - &\red - &\red - &\red - \\
 employer &\green 0.023 &\green - &\green - &\green 0.942 &\green - &\green 3.167 \\
 member of &\green 0.152 &\green - &\green - &\green - &\green - &\green 3.352 \\
 instance of &\red - &\red - &\red - &\green - &\green 1.239 &\green - \\
 developer &\red - &\red - &\red - &\green - &\green 0.501 &\green - \\
 language of work or name &\red - &\red - &\red - &\green - &\green - &\green 3.823 \\
 country of origin &\red - & \red - &\red - &\green 0.298 &\green - &\green 1.591 \\
 original language of work &\red - &\red - &\red - &\green 0.416 &\green - &\green 2.457 \\
 owned by &\red - &\red - &\red - &\green 1.293 &\green - &\green - \\
\bottomrule
\end{tabular}
\caption{\textbf{Per-relation breakdown} of three classes of facts,  categorized by their appearance in the generated paragraphs produced by InstructGPT and FLAN-T5, is presented to evaluate the effectiveness of the \textit{entropy} metric in distinguishing between facts across these classes. Relations in which our metric demonstrates effective differentiation between different fact classes are highlighted in green.}
\label{tab:factual-ent}
\end{table*}

\end{document}